\documentclass[runningheads]{llncs}
\usepackage{graphicx}
\usepackage{hyperref}
\usepackage{algpseudocode}
\begin{document}
\title{Faster object tracking pipeline for real time tracking \thanks{Supported by Infocusp Innovations Private Limited.}}


%
%
\author{Parthesh Soni\inst{1,2}\orcidID{0000-0002-8416-4779} \and
Falak Shah\inst{1}\orcidID{0000-0003-1126-5840} \and
Nisarg Vyas\inst{1}}


%
\institute{Infocusp Innovations Private Limited \\
\email{\{parthesh, falak, nisarg\}@infocusp.in}\\
\url{http://infocusp.in/}\\
}
\maketitle              
\begin{abstract}
Multi-object tracking (MOT) is a challenging practical problem for vision based applications. Most recent approaches for MOT use precomputed detections from models such as Faster RCNN, performing fine-tuning of bounding boxes and association in subsequent phases. However, this is not suitable for actual industrial applications due to unavailability of detections upfront. In their recent work, Wang et al. proposed a tracking pipeline that uses a Joint detection and embedding model and performs target localization and association in realtime. Upon investigating the tracking by detection paradigm, we find that the tracking pipeline can be made faster by performing localization and association tasks parallely with model prediction. This, and other computational optimizations such as using mixed precision model and performing batchwise detection result in a speed-up of the tracking pipeline by 57.8\% (19 FPS to 30 FPS) on FullHD resolution. Moreover, the speed is independent of the object density in image sequence. The main contribution of this paper is showcasing a generic pipeline which can be used to speed up detection based object tracking methods. We also reviewed different batch sizes for optimal performance, taking into consideration GPU memory usage and speed.

\end{abstract}
\section{Introduction}
Multi object tracking systems are the basic building blocks for tasks related to surveillance \cite{xu2018real,xie2004multi}, robotics \cite{bai2016multi,schmitt2002watch}, autonomous vehicles \cite{rangesh2019no,ess2010object} and activity recognition \cite{izadinia2013multi}. Tracking is a challenging problem owing to the associated problems such as abrupt changes in target location, illumination changes, occlusions and reflections. In applications like autonomous vehicles, speed of the tracking system is crucial. Even seconds of delay can prove to be fatal in some cases. This is why state of the art tracking systems aim to combine good performance and faster implementation. 

\subsection{Joint Detection and Embedding approach}
 Majority of the trackers suggested in computer vision literature that work in realtime assume availability of prior detections for the frames in the video sequence \cite{bewley2016simple,wojke2017simple}. This makes them unsuited for use in real time applications due to unavailability of the detections upfront. However, in \cite{wang2019realtime}, Wang et al. suggest an approach which does not require prior detections. A Joint Detections and Embeddings (JDE) model is proposed, where the model gives detections as well as appearance embeddings for the detections in a single forward pass. Appearance embeddings can be thought of as compressed information about the object appearing in the corresponding detection. The post-processing part performs association of detected objects across frames using similarity of these embeddings and Kalman filter. The Kalman Filter uses detections obtained by the model as measurements, and predicts the location of object in current frame using locations in past frames. In this paper, we build further on this approach by Wang et al. to achieve more speedup.

\begin{figure}[h!]
\centering{
            \includegraphics[width=.32\textwidth]{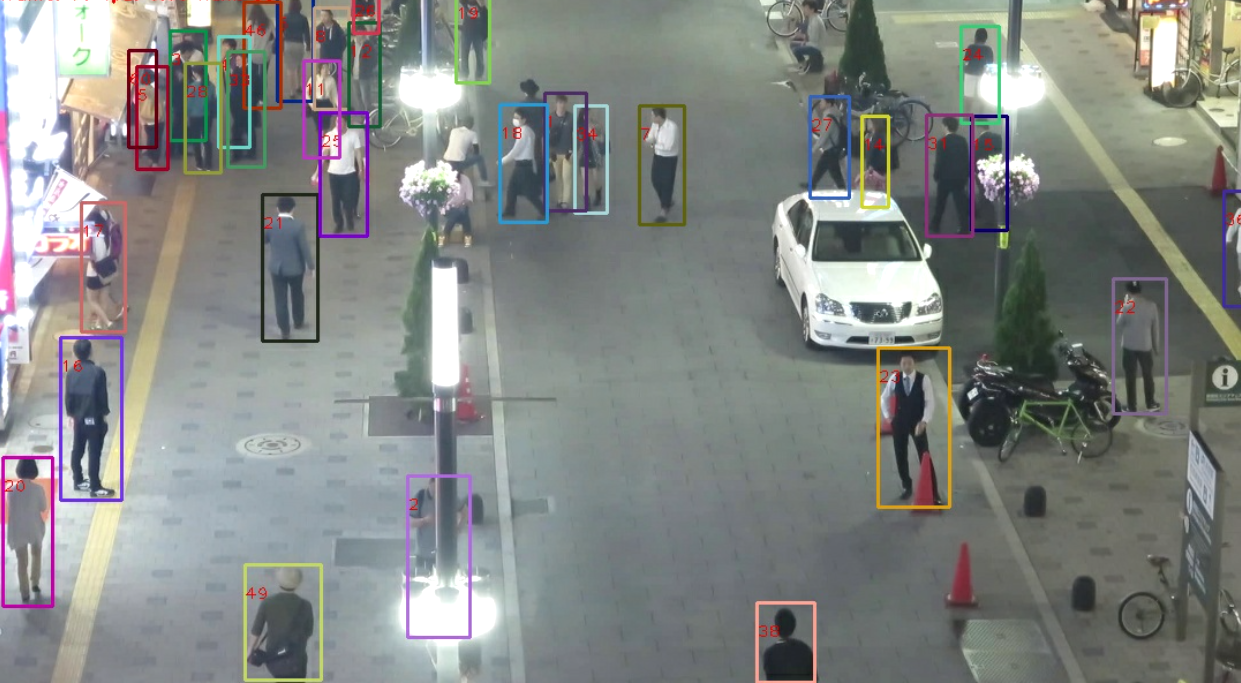}
            \includegraphics[width=.32\textwidth]{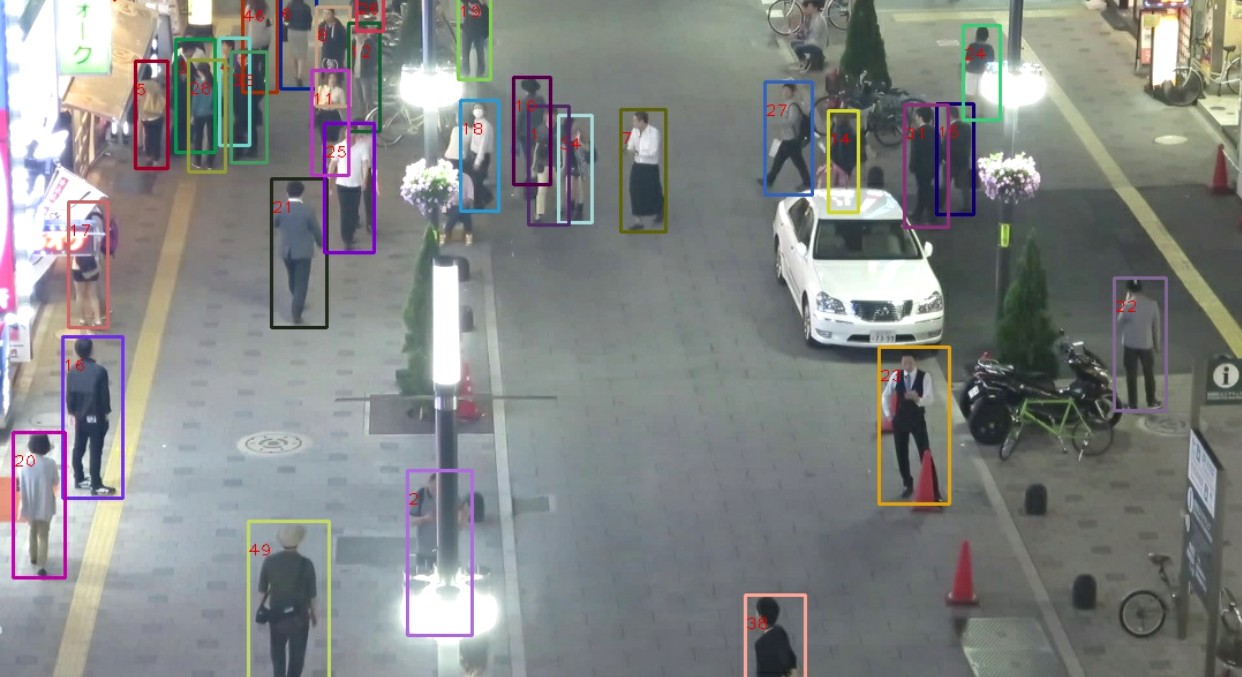}
            \includegraphics[width=.32\textwidth]{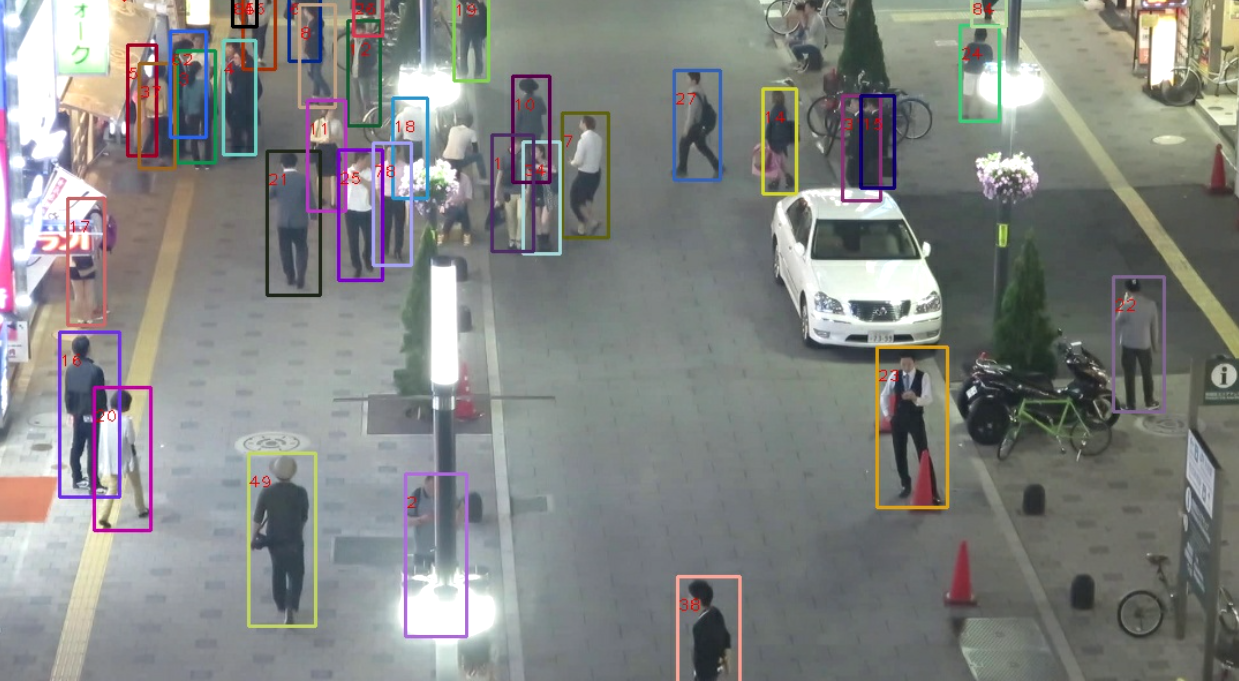}\\
            \includegraphics[width=.32\textwidth]{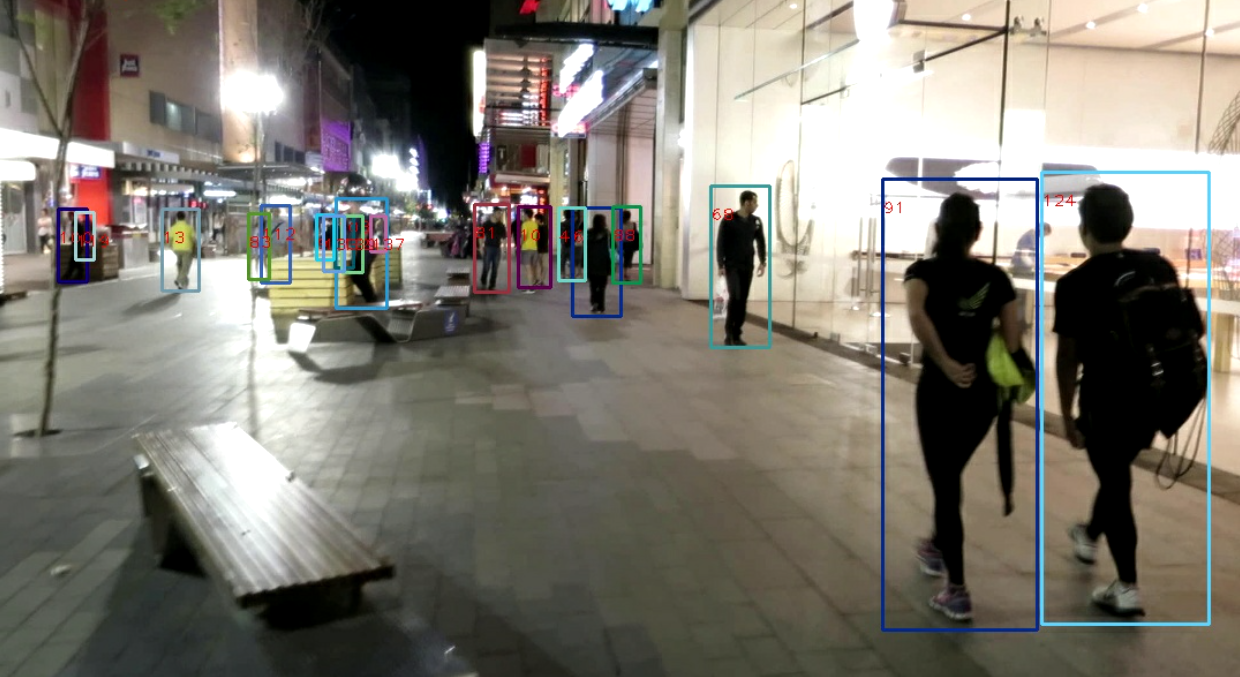}
            \includegraphics[width=.32\textwidth]{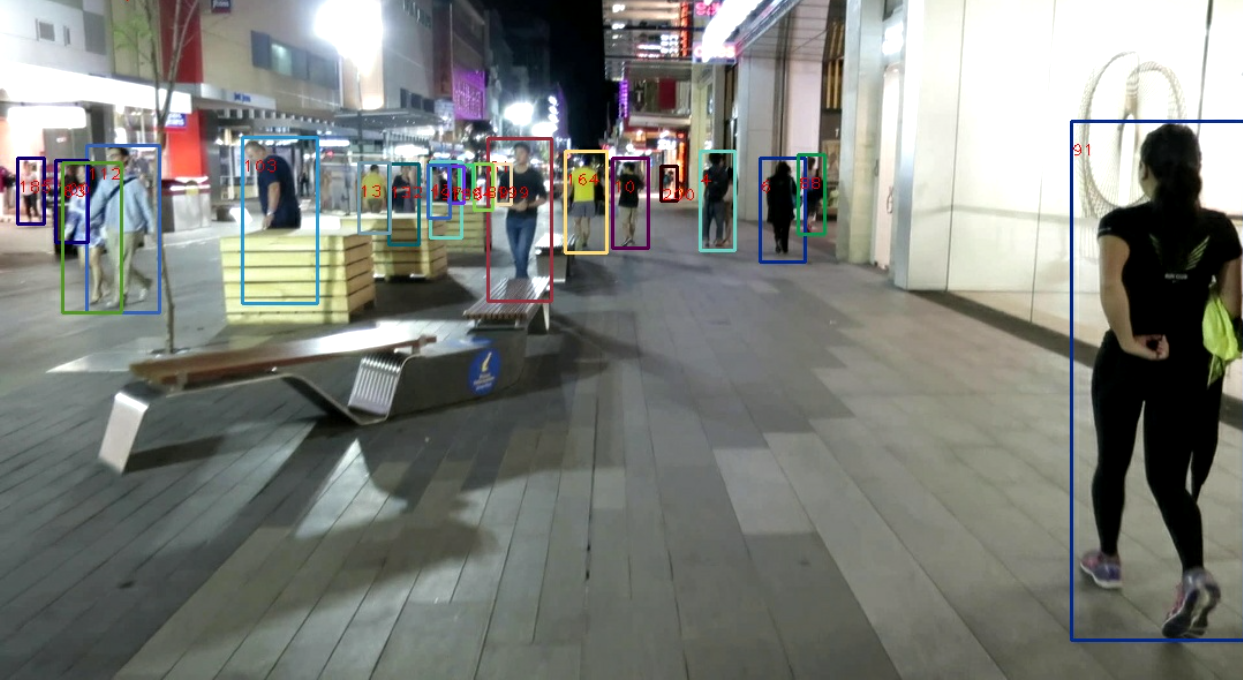}
            \includegraphics[width=.32\textwidth]{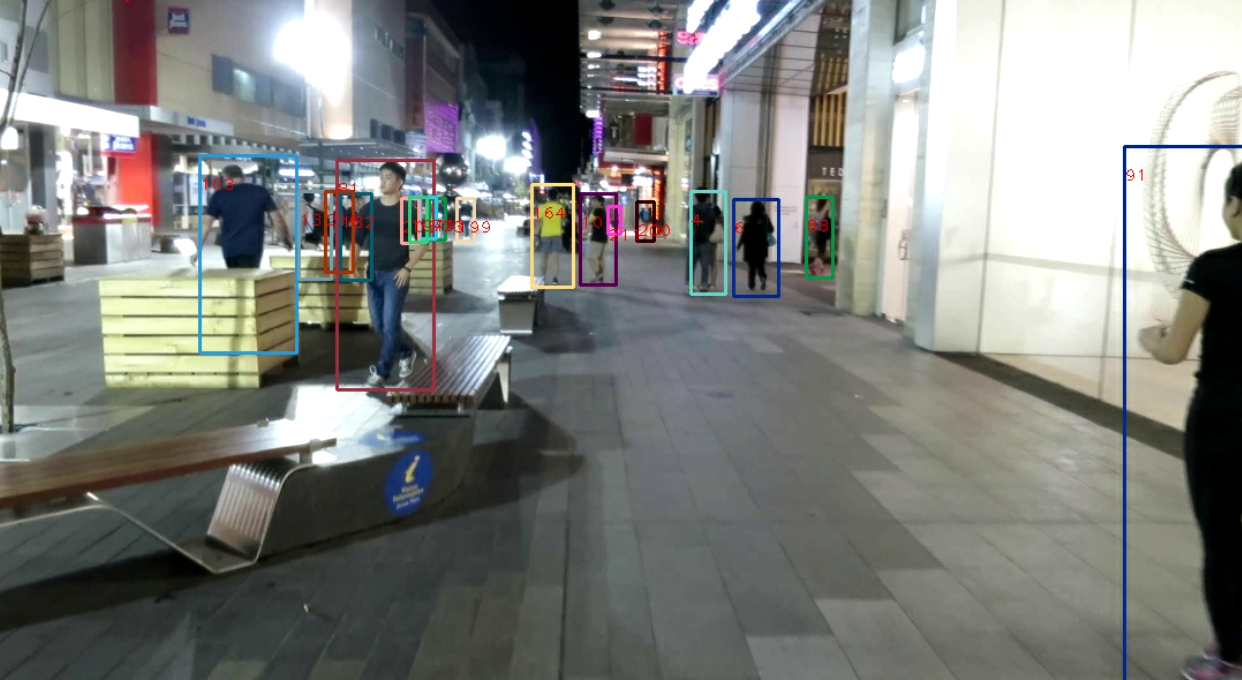}
            \caption{Result of our tracker on MOT16 video sequences}\label{img:results}}
\end{figure}

We performed time profiling of \cite{wang2019realtime} and found that the detection by model was the most time consuming portion. It accounts for almost 80\% of the inference time and the post-processing step takes 20\%. Although the model works at 19 FPS on FullHD resolution, we found aspects that can be optimized further. The main contribution of this paper is to suggest optimizations, namely: model compression, batch wise processing and concurrent post processing for boosting the performance of the JDE tracker. These optimizations can in general be applied to any trackers following detection by tracking paradigm. With these in place, we gain 57.8\% speedup and the tracking speed is independent of the object density in the image sequences. Some sample results of our tracker are shown in Figure \ref{img:results}.

The rest of the paper is organized as follows:
In section \ref{sec:rpw}, we survey past work done in the domain of object tracking. Section \ref{sec:tbd} describes the tracking algorithm which is our main focus along with model architecture details and post processing steps. In order to understand mixed precision used during inference time, readers can refer Section \ref{sec:mop}. Details of our proposed pipeline are mentioned there. Section \ref{sec:res} of the paper shows comparison of results.

\section{Relevant past work}\label{sec:rpw}
Object trackers are majorly divided into two categories: single object trackers and multi-object trackers. Single object trackers include MD-Net \cite{nam2016learning} that requires domain specific training, followed by online updates and operates at 10 frames per second (FPS). Recurrent YOLO (ROLO) model \cite{ning2017spatially} is another tracker for single object tracking that uses a combination of high-level visual features produced by Deep CNN and Long Short Term Memory (LSTM) to get accurate region information.

State of the art multi-object tracking approaches are leaning rapidly towards tracking by detection paradigm \cite{yu2016poi,milan2016mot16,wen2014multiple,zamir2012gmcp,kim2015multiple,bewley2016simple}. They comprise of two main steps: performing object detection using a CNN based deep learning model and associating the objects in those detections across frames. 


These trackers generally utilize two techniques for association:  association by optimization on graphs \cite{wen2014multiple,zhang2008global}, and using neural networks \cite{baser2019fantrack,yoon2019data}.
Convolutional Neural Networks (CNNs), Fully connected networks (FCNs) and Long Short Term Memory (LSTM) networks are common neural network architectures whereas min-cost flow network, undirected hierarchical relation hypergraph are common graph optimization approaches used for the association task. In these approaches, detections are given as input, while the system is responsible for associating those detections across frames. 


One other category has been proposed that does not involve the association phase. \cite{bergmann2019tracking} accomplishes the task without any specific training related to tracking. This tracker makes extensive use of the bounding box regressor, regressing on the objects which are assumed to have moved only little in the next frame. This approach requires additional reidentification networks like Siamese network for proper functioning which reduces the speed of the overall system.

 The tracking systems SORT \cite{bewley2016simple} and Deep SORT \cite{wojke2017simple} achieve very high tracking speed, but without accounting for the time taken to perform detection. In industrial or real time applications, detections would not be available prior in the video feed, thus rendering these systems incapable for pratical use.

\section{Tracking by Detection}\label{sec:tbd}

The methods using the "tracking by detection" paradigm have the following main phases in the pipeline:
\begin{enumerate}
    \item  Detect objects in the frame.
    \item  Estimate the location of the object in current frame based on location in the previous frames.
    \item  Combine information about the detections and estimated location to locate the object, based on similarity \cite{murray2017real}.
\end{enumerate}

\textbf{Detection}: In this step, the object detection model predicts the location of the object in a given frame. The input to the object detection models is an image and the outputs are the coordinates of the bounding boxes and their confidence scores. If the object detection is performed for multiple classes, labels are also provided for getting the class information of the detection. Common models used for obtaining detections are Faster-RCNN (used in \cite{bewley2016simple}, \cite{wojke2017simple}),  R-FCN (in \cite{feichtenhofer2017detect}) and YOLO (in \cite{peixoto2019mice}). In case of JDE, the model also outputs appearance embeddings.

\textbf{Location Estimation}: This stage provides the estimate of target object's location in the current frame based on its observed or predicted state (such as location, appearance information and velocity) obtained from the previous frames. Methods like Kalman filter (used in \cite{bewley2016simple,wojke2017simple,wang2019realtime}, optical flow (in \cite{xiang2015learning}) or particle filter (in \cite{sanchez2016online,yu2016poi,smith2005using}) are used for estimating the current location or state of the object. Our pipeline uses Kalman filter for the prediction task.

\textbf{Association}: In this task, detections are assigned to tracklets, based on the similarity between detections and predictions \cite{murray2017real}. Here, tracklet refers to a fragment of track or path followed by the target object across frames. A cost matrix is computed that measures the distance between the features which are obtained by embeddings of the detections in the current frame and that of the existing tracklets. Hungarian algorithm \cite{kuhn1955hungarian} is a popular method used for this association task. It assigns cost to every tracklet detection pair, and then assignment is selected such that the overall cost is minimum. \cite{wang2019realtime,bewley2016simple,wojke2017simple,luetteke2012implementation} use Hungarian algorithm for association task.

Our main focus in this paper is the JDE model that gives detections and corresponding embeddings in a single forward pass. The objective of the training would be to train the model in such a way that the coordinates of the bounding box become as near to the ground truth as possible. Also, the embeddings of object which is in the current frame should be as near as possible to the embeddings of same object in previous frames, distance being measured by Euclidean distance. Also, the distance of embeddings of different objects should be farther. \cite{koller1994towards}.


\subsection{Model architecture}
The model architecture mainly consists of two parts: Darknet-53 as backbone network, with other layers added on top for detections and embeddings. Detection is performed in same manner as in Yolov3 \cite{redmon2018yolov3}.

Darknet weights are pre-trained on ImageNet, and act as a general feature extractor. The JDE model is then further trained on six publicly available datasets: ETH dataset, CityPersons \cite{Shanshan2017CVPR} dataset, CalTech dataset, MOT-16 \cite{MOT16} dataset, CUHK-SYSU dataset, PRW \cite{zheng2017person} dataset. 

\begin{figure}
    \centering
    \includegraphics[scale=0.3]{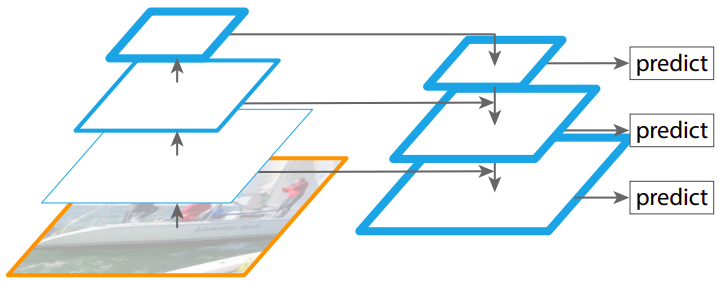}
    \caption{Feature Pyramid Network, which is used in the implementation. Figure reprinted from \cite{lin2017feature}}
    \label{fig:fpn}
\end{figure}

 Feature Pyramid Network (FPN) \cite{lin2017feature} is used in JDE model, that works well in detection of objects that have large variations in size. With FPN, predictions are obtained at various scales. Hence, they can easily detect varied sized objects. As shown in Figure \ref{fig:fpn}, different feature maps are employed for prediction. The input passes through the network, and its features are extracted at different stages in the network, which are used by prediction heads. Prediction heads are layers which generate the final output of the model. Higher level features are upsampled and fused with the lower level features to improve the prediction at lower level. In JDE, we get output from three different layers in the architecture.

\subsection{Post processing steps}
From the detections, the first step is to keep detections which have objectness score greater than a threshold. Then, non-max supression is applied to further filter the detections. Kalman filter is used to estimate the updated locations of past detections using the state. The association is then performed using the hungarian alorithm on the embeddings. More details of the post processing steps are mentioned in Appendix \ref{sec:appa}.

\section{Optimization of tracking pipeline}\label{sec:mop}

In the original work \cite{wang2019realtime}, the author used serialized approach for tracking where detection is followed by post processing steps mentioned above. We converted the full-precision model to mixed-precision, that resulted in speedup without significant loss in the performance. The original implementation performed inference on one sample at a time. We are also processing images in batches which improves the speed, even with serialized post processing. Lastly, we found that the model prediction is independent of the post processing and both can run parallely. A separate daemon process can be started for the post processing with the prediction writing output to a queue from which post processing picks the inputs. This gives speedup and makes the performance independent of the object density as the post processing is happening concurrently with prediction. Details of all three steps are given below.



\subsection{Model downsizing using mixed precision}
Mixed precision refers to process of reducing the precision level of majority of the operations to half (i.e. to float16 from float32). We used Apex \cite{apex}, which converts all the numerically safe operations to float16 precision while retaining the batch norm and other similar operations to float32. 
Using mixed-precision helps in boosting the speed at training and inference time and also reduce the GPU memory requirement. We were able to gain speed from 19 fps to 23 fps on Nvidia Titan X without any drop in quantitative performance. Interested readers please refer to \cite{micikevicius2017mixed} for more information on mixed precision.


\subsection{Batch Processing}
Batch processing refers to simultaneous processing of a batch of input images in a network, i.e, a batch of data is fed into the network and the network gives output for all inputted data at once.  We tried different batch sizes, from 1 to 10 and noted different trends of speed for each batch-size update. Detailed comparison is shown in the results section.

\subsection{Parallel post processing}
We perfomed time analysis of each section of the codebase. It showed that the most time consuming part of the pipeline was the forward pass through the model itself. We discovered that the functionality of model is independent from the post processing part and the model forward pass can be  run parallelly with the post processing part. In our implementation, the main process only handles forward pass, not waiting for the post processing of the detections and embeddings, hence saving time. 

\subsection{Updated tracking pipeline}
Our tracking pipeline working in near realtime scenario is as follows:
\begin{enumerate}
    \item Video Capturing:  A process that is capturing images from device/ reading from disk and writing out to a shared queue for the model to read from.
    \item Model prediction: The model predicts the detections and embeddings of the given frames in a batch.  The output from the model is stored in another queue, which is checked regularly by the post processing function. 


    \item Post-processing (Seperate daemon process): This method is responsible for the post processing steps detailed in Appendix \ref{sec:appa}.  It works in parallel to the model forward pass as a separate process and uses the predictions that the model writes out to the queue. 
\end{enumerate}

\begin{figure}
\makebox[\textwidth][c]{
    \includegraphics[width=1.2\textwidth]{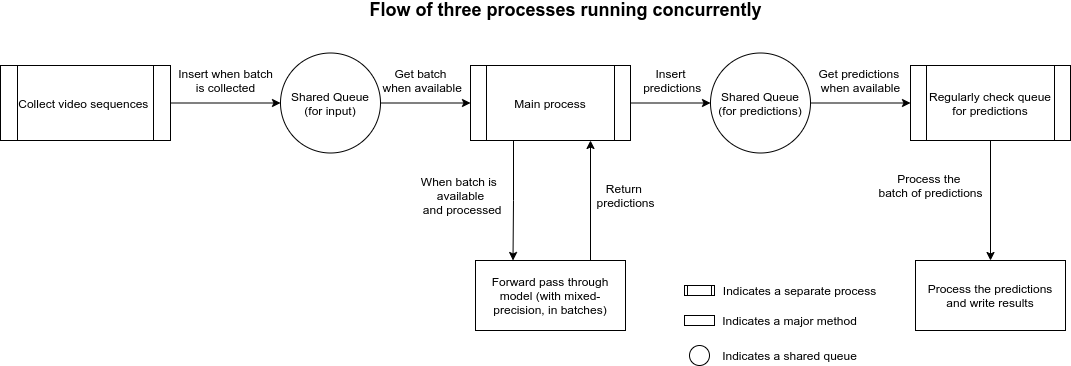}}
    \caption{Graphical representation of new tracking pipeline.}
    \label{fig:flow}
\end{figure}

Figure \ref{fig:flow} shows the flow graphically.

\section{Results} \label{sec:res}
With the above optimizations applied to the model, we encountered almost no change in performance metrics compared to the original approach. With similar performance, we get significant increase in speed and decrease in GPU memory foot-print (with batch-size 1). The metrics computed for MOT16 dataset are compared in Table \ref{tab1}. Refer \cite{py-motmetrics} for more information on the evaluation metrics. 

\begin{table}
\caption{Comparison of quantitative metrics.}\label{tab1}
\begin{tabular}{|l|c|c|c|c|c|c|c|c|c|c|c|c|c|c|}
\hline
 & IDF1  & IDP  & IDR & Rcll & Prcn& MT & PT & ML &  FP &   FN & IDs & FM & MOTA & MOTP \\
\hline
Original & 68.9\% & 74.4\% & 64.1\% & 80.3\% & 93.1\%& 324 & 165 & 28 & 6593 & 21788 & 1312 & 2345 & 73.1\% & 0.183 \\
\hline
Our & 73.1\% & 78.7\% & 68.2\% & 80.8\% & 93.2\% & 324 & 165 & 28 & 6556 & 21252 & 1229 & 2702 & 73.7\% & 0.180 \\
\hline
\end{tabular}
\end{table}



\begin{figure}[h!]
    \centering
    \includegraphics[width=0.8\textwidth]{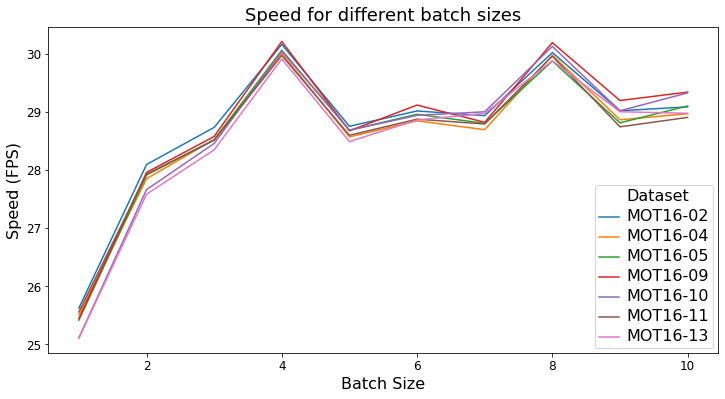}
    \caption{The graphs show inference speed in FPS measured for every batch-size for all sequences}
    \label{fig:fps}
\end{figure}

\begin{figure}[h!]
    \centering
    \includegraphics[width=0.8\textwidth]{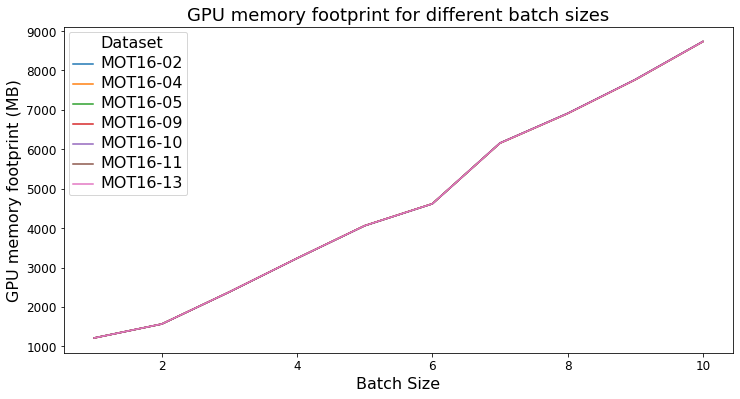}
    \caption{The graphs show GPU memory footprint for every batch-size for all sequences in MOT16}
    \label{fig:GPU}
\end{figure}

We analysed speed and GPU memory utilization for varying batch sizes for all the video sequences in MOT16 dataset which is shown in Figure \ref{fig:fps} and Figure \ref{fig:GPU}. All experiments were performed on Titan X GPU.

 We compared performance for batch-sizes of 1 to 10. We find that the optimal batch size which gives a good tradeoff between speed and memory footprint is 4. In case of severe memory constraints, we show that batch size of 1 can give speed boost of about 6 FPS (19 FPS to 25.5 FPS), in the same time using lesser memory footprint(reduced from 2171 MB to 1535 MB). Also interesting to note is the peaking of performance at multiples of 4 owing to 4 concurrent processors on GPU.
 

\subsection{Speed analysis}
\begin{figure}
\centering{\includegraphics[width=0.9\textwidth]{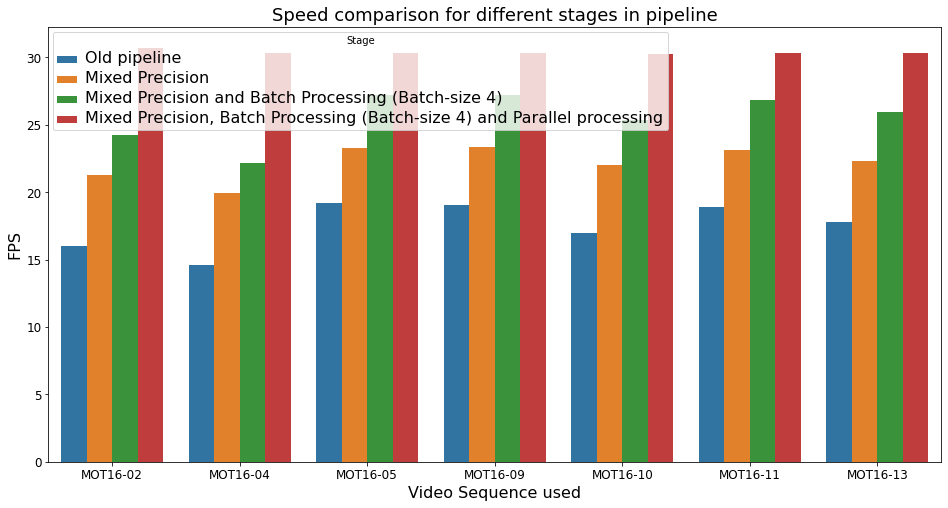}}
\caption{Comparison of speed with additional updates to the pipeline} \label{fig:speed}
\end{figure}

Table \ref{tab2} compares the speed for different approaches. First column indicates the video sequence used. All the sequences are taken from the training set of MOT16 challenge, which are in FullHD resolution. The speeds are also shown in Figure \ref{fig:speed}

\begin{table}
\caption{Speed(in FPS) for different pipelines.}\label{tab2}
\centering{
\begin{tabular}{|l|c|c|c|c|}
\hline
Video Sequence  & OP  & MP & MP and BW & MP, BW and PP \\
\hline
MOT16-02 & 15.99 & 21.30 & 24.22 & 30.73 \\
\hline
MOT16-04 & 14.61 & 19.94 & 22.16 & 30.33\\
\hline
MOT16-05 & 19.20 & 23.31 & 27.18 & 30.30\\
\hline
MOT16-09 & 19.06 & 23.32 & 27.20 & 30.30\\
\hline
MOT16-10 & 16.95 & 22.02 & 25.37 & 30.26\\
\hline
MOT16-11 & 18.92 & 23.13 & 26.81 & 30.35\\
\hline
MOT16-13 & 17.81 & 22.33 & 25.98 & 30.33\\
\hline
\end{tabular}}
\end{table}
The updates to pipeline that are mentioned in table are as below
\begin{itemize}
    \item OP: Original pipeline from \cite{wang2019realtime}
    \item MP: Model changed to use mixed precision 
    \item MP and BW: Mixed precision model and batchwise inference
    \item MP, BW and PP: Parallel post processing in addition to the optimizations above. Note that with this update, the speed is almost independent of the video sequences
\end{itemize}

    
    
    

The code has been uploaded \href{https://anonymous.4open.science/r/d3922805-d11d-4a5c-ad1f-6e11c209c8dd/}{here} for anonymity and will be updated to GitHub post review.

\section{Conclusion}
In this paper, we proposed a new generic pipeline for tracking systems following the tracking by detection paradigm. We extend the tracking system of \cite{wang2019realtime} with our pipeline and demonstrate speedup. We employ parallel processing of independent steps, along with mixed-precision model and batch-wise processing. The performance metrics of the system remains almost same compared to the original pipeline. Hence, realtime speed is obtained without sacrificing the performance. Also, the performance is now independent of the object density within the frames.
%
%
%
\bibliographystyle{splncs04}
\bibliography{bibliography}

\section{Appendix}
\appendix
\section{Post Processing steps}\label{sec:appa}
The post-processing steps are as follows: 
\begin{itemize}

    \item The JDE model returns a tensor of dimension [n, num\_detections, embeddings\_len (518 in our case)]. Here 'n' is the batch size used for the system. The second dimension is the number of detections returned by the model. For third dimension, the first 4 values indicate the location of the object, followed by object confidence score and class score. The remaining 512 values are the embeddings of the detection. 

    \item Filter detections based on objectness score: If object score is lesser than threshold confidence, the detection is discarded.

    \item Non-max supression: Sort detections based on object confidence score and discard detections having IOU above threshold with highest scored detection.


    \item Predict the locations of the detections of the tracks which are active, till last frame using kalman filter. 

    \item Cosine distance is calculated between the embeddings obtained for the detections in current frame and that in the tracks which are active or lost for some frames. 

    \item If the distance between the predicted location of object in track and current detections is greater than a threshold, they are rejected and 'infinte' cost is assigned in corresponding cell of the cosine distance matrix.

    \item After the cost matrix is refined, hungarian algorithm is used for assignment of detections to tracks. Kalman filter state is updated. Assignments are rejected if the cost of the pairs is greater than a threshold. If a detection gets assigned to a track, the track is marked as active.


    \item The tracks which are not assigned any detections are marked as lost.


    \item For all the lost tracks, if they are lost for frames greater than given threshold, they are removed.
\end{itemize}

\end{document}